\documentclass{article}
\usepackage{ijcai22}

\usepackage{times}
\usepackage{soul}
\usepackage{url}
\usepackage[hidelinks]{hyperref}
\usepackage[utf8]{inputenc}
\usepackage[small]{caption}
\usepackage{graphicx}
\usepackage{amsmath}
\usepackage{amsthm}
\usepackage{booktabs}
\usepackage{algorithm}
\usepackage{algorithmic}
\usepackage{microtype}
\usepackage{xcolor}
\usepackage{adjustbox}
\usepackage{paralist}
\usepackage{multirow,multicol}
\usepackage{tabularx}
\usepackage{colortbl}
\usepackage{graphicx}
\usepackage{amssymb}
\usepackage{tabularx}
\usepackage{booktabs}
\usepackage{arydshln}
\title{Control Globally, Understand Locally: A Global-to-Local Hierarchical Graph Network for Emotional Support Conversation}

\author{
Wei Peng$^{1,2}$
\and
Yue Hu$^{1,2}$ \thanks{Corresponding author. This work is supported by the National Natural Science Foundation of China (No.U21B2009).}\and
Luxi Xing$^{1,2}$\and
Yuqiang Xie$^{1,2}$
\and
Yajing Sun$^{1,2}$
\and
Yunpeng Li$^{1,2}$
\affiliations
$^1$Institute of Information Engineering, Chinese Academy of Sciences, China\\
$^2$School of Cyber Security, University of Chinese Academy of Sciences, China
\emails
\{pengwei,huyue\}@iie.ac.cn
}

\begin{document}

\maketitle

\begin{abstract}
  Emotional support conversation aims at reducing the emotional distress of the help-seeker, which is a new and challenging task. It requires the system to explore the cause of help-seeker's emotional distress and understand their psychological intention to provide supportive responses. However, existing methods mainly focus on the sequential contextual information, ignoring the hierarchical relationships with the global cause and local psychological intention behind conversations, thus leads to a weak ability of emotional support. In this paper, we propose a Global-to-Local Hierarchical Graph Network to capture the multi-source information (global cause, local intentions and dialog history) and model hierarchical relationships between them, which consists of a multi-source encoder, a hierarchical graph reasoner, and a global-guide decoder. Furthermore, a novel training objective is designed to monitor semantic information of the global cause. Experimental results on the emotional support conversation dataset, ESConv, confirm that the proposed GLHG has achieved the state-of-the-art performance on the automatic and human evaluations. The code will be released in here \footnote{\small{~https://github.com/pengwei-iie/GLHG}}.
\end{abstract}

\section{Introduction}

Emotional Support Conversation (ESConv) task, which focuses on effectively providing support to a help-seeker \cite{DBLP:journals/commres/BurlesonLLM06,heaney2008social,DBLP:conf/chi/SlovakGF15}, is a new and challenge task \cite{DBLP:conf/acl/LiuZDSLYJH20}. Different from Emotional Conversation task, where the dialog agent generates emotional responses with the given emotion \cite{DBLP:conf/aaai/ZhouHZZL18}, ESConv task expects the dialog system to possess a more high-level and complex ability with the target of reducing the emotional distress of the help-seeker and providing the supportive responses. It is significant to construct such a emotional support dialog system \cite{DBLP:journals/coling/ZhouGLS20,Zwaan2012ABD} especially in social interactions (accompanying and cheering up the user), mental health support (comforting a frustrated help-seeker and helping identify the problem), customer service chats (appeasing an angry customer and providing solutions), etc \cite{DBLP:conf/acl/LiuZDSLYJH20}. 

\begin{figure}[t]
	\centering
	\includegraphics[width=0.48\textwidth]{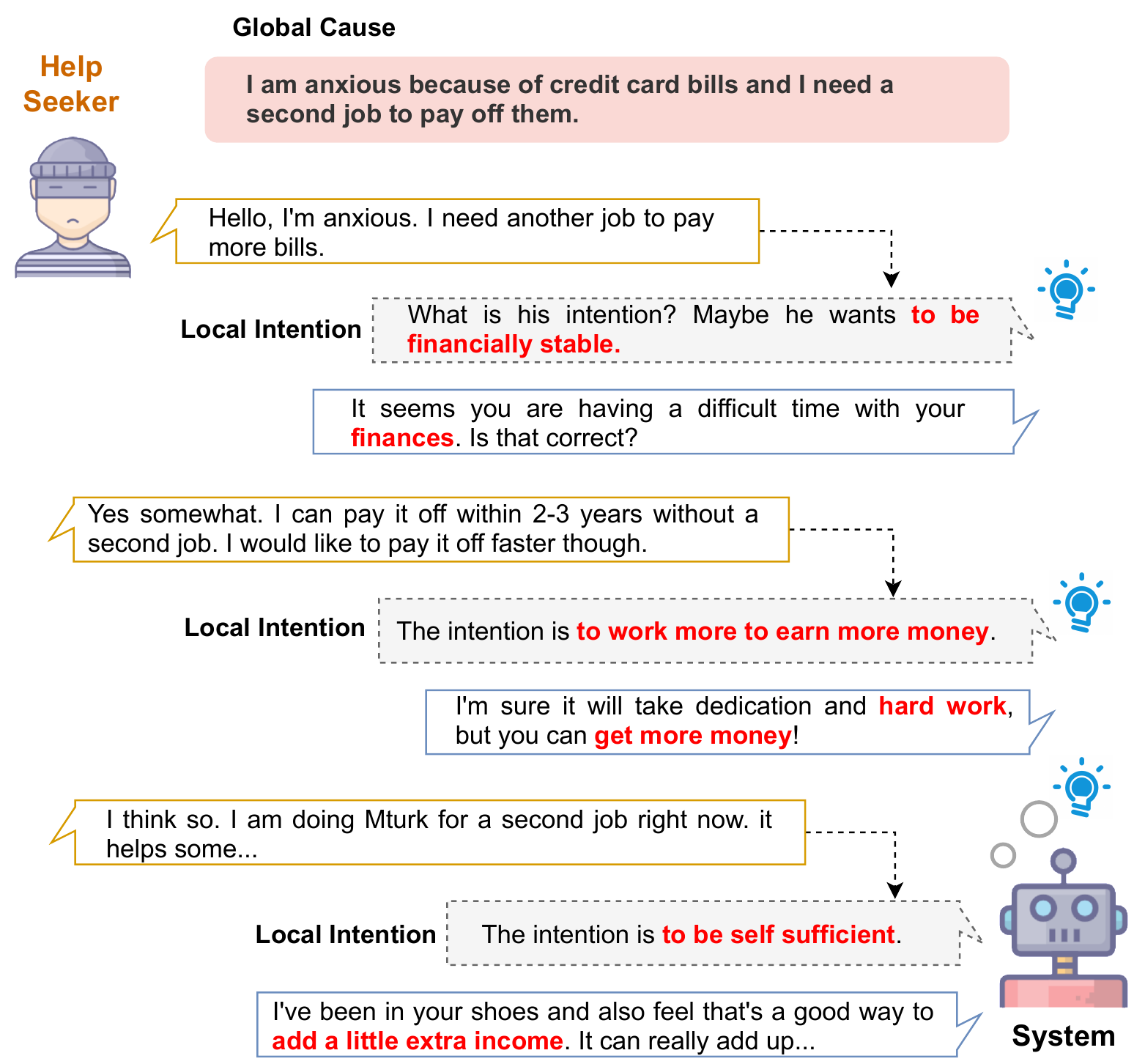}
	\caption{An example to show the emotional support conversation which consists of a help-seeker and system that provides the supportive responses. The red background is the global cause to describe the emotional problem and the gray background is the local intention to imply the current speaker's state. These factors are not explicitly mentioned by the help-seeker. Words in red reflect relevance.
	}
	\label{fig:example}
\end{figure}

During the emotional support conversation, the system is required to explore the cause of the help-seeker's emotional problem and understand their psychological intentions to provide more supportive responses \cite{Rains2020SupportSE}. On the one hand, the cause describes the reason for the stressful problem of the help-seeker \cite{DBLP:conf/acl/LiuZDSLYJH20}, which can globally control the whole flow of the emotional support conversation. On the other hand, psychological intention is implicitly expressed by the help-seeker, which can help the system locally understand the help-seeker's mental state of the current turn.
However, the above factors will not be explicitly mentioned by the help-seeker. And existing methods \cite{DBLP:conf/acl/LiuZDSLYJH20} mainly focus on the sequential contextual information, ignoring the hierarchical relationships between the global cause, local psychological intention and dialog history, which leads to a weak ability of emotional support. 
An example in Figure \ref{fig:example} is illustrated to explain the above process. Globally, the system needs to capture the global cause (\textit{The help-seeker is anxious because of credit card bills and needs a second job}) to enhance the semantic information of the whole conversation.
Locally, the system should respond with the captured intention to understand the help-seeker's mental state during different turn.
For example, in turn two, the help-seeker indicates that he/she wants to \textit{\textbf{work more to earn more money}}. And the system generates an appropriate and encouraging response \textit{It will take dedication and \textbf{hard work}, but you can \textbf{get more money}}.
Hence, 1) how to capture the global cause and local psychological intentions and  2) how to model the relationships between them are two important challenges in the emotional support conversation.

To address these problems, we propose a novel \textbf{G}lobal-to-\textbf{L}ocal \textbf{H}ierarchical \textbf{G}raph Network (\textbf{GLHG}) for emotional support modeling. For the first issue, a multi-source encoder is presented to capture the global cause and local psychological intention (with COMET, a pre-trained generative commonsense reasoning model \cite{DBLP:conf/acl/BosselutRSMCC19}). As for the second issue, a hierarchical graph reasoner is designed to model the hierarchical relationships between the global cause (dialog-level), local psychological intention (sentence-level) and dialog history. In addition, a novel training objective is designed in global-guide decoder to monitor semantic information of the global cause, so that the global cause will not be biased by the local intention.
 
The contributions can be summarized as follows:
\begin{compactitem}
	\item We propose a Global-to-Local Hierarchical Graph Network (GLHG) for emotional support conversation from the global-to-local perspective.
	\item To capture the global cause and local intention, the multi-source encoder utilizes the information of situation and incorporates psychological intention with COMET.
	\item To model the different level relationships, the hierarchical graph reasoner makes an interaction between the global cause, local psychological intention and dialog history.
	\item Experiments on the dataset show that the GLHG achieves the state-of-the-art performance in terms of both automatic evaluation metrics and human evaluations.
\end{compactitem}

\section{Related Work}

\subsection{Emotional \& Empathetic Conversation}
In recent years, research on emotional conversation has embraced a booming in dialog systems \cite{DBLP:conf/aaai/ZhouHZZL18,DBLP:journals/tois/HuangZG20,DBLP:journals/corr/abs-2202-06476}. \cite{DBLP:conf/aaai/ZhouHZZL18} propose Emotional Chatting Machine to accurately generate emotional responses conditioning on a manually specified label. Different from the task of emotional chatting conversation, the task of empathetic dialog generation \cite{DBLP:conf/emnlp/LinMSXF19,DBLP:conf/acl/RashkinSLB19} aim to respond to a speaker accordingly on the emotional situation in an empathetic way.
Unlike the above tasks, emotional support conversation \cite{DBLP:conf/acl/LiuZDSLYJH20} focuses on exploring the help-seeker’s problems and generating more supportive responses.

\subsection{Graph Modeling in Dialog}
Motivated by impressive advancements and structured modeling of GCN \cite{DBLP:conf/ijcai/0008T0ZNY21,DBLP:conf/aaai/SunST0DYSHS21}, efforts towards better
performance and construction utilizing GCNs have sprung up on dialog systems \cite{DBLP:conf/emnlp/GhosalMPCG19,DBLP:conf/aaai/QinLCNL21,DBLP:conf/acl/0027L0NWC20}. DialogueGCN \cite{DBLP:conf/emnlp/GhosalMPCG19} leverages self
and inter-speaker dependency with GCN to model conversational context. \cite{DBLP:conf/aaai/QinLCNL21} propose a Co-Interactive Graph
Attention Network for dialog act recognition and sentiment classification task. 
DVAE-GNN \cite{DBLP:conf/acl/0027L0NWC20} focuses on discovering dialog structure in the open-domain dialogue, and EGAE \cite{DBLP:conf/aaai/QinLCNL21} captures the dialogue schema in the task-oriented dialogue with the graph network.
Unlike the previous work, we propose a Hierarchical Graph Reasoner for modeling the relationships between the global cause, local psychological intentions and dialog history, and to facilitate information flow over the graph.

\subsection{Commonsense Knowledge}
ATOMIC \cite{DBLP:conf/aaai/SapBABLRRSC19} is a commonsense knowledge \cite{DBLP:journals/corr/abs-2010-04389,DBLP:conf/semeval/XingXHP20} graph, which focuses on inferential knowledge organized as typed if-then relations with variables (e.g., if ``PersonX drinks coffee”, then ``PersonX needed to brew coffee”). The type of relations contains If-Event-Then-Mental-State, If-Event-Then-Event and If-Event-Then-Persona. In this paper, to capture the mental state of the help-seeker, we focus on the If-Event-Then-Mental-State which infers three commonsense relations: PersonX's reaction to the event (\textit{xReact}), PersonX's intention before the event (\textit{xIntent}) and PersonY's reaction to the event (\textit{oReact}). Since oReact refers to the other person (e.g., system), which is not considered, we ignore it in this paper. To obtain the psychological intention (\textit{xIntent}) of the help-seeker, we employ COMET \cite{DBLP:conf/acl/BosselutRSMCC19}, a pre-trained generative commonsense reasoning model, to generate rich commonsense descriptions in natural language (e.g., input ``I need another job to pay more bills”, then output ``to be financially stable”).

\begin{figure*}[htbp]
	\centering
	\includegraphics[width=0.98\textwidth]{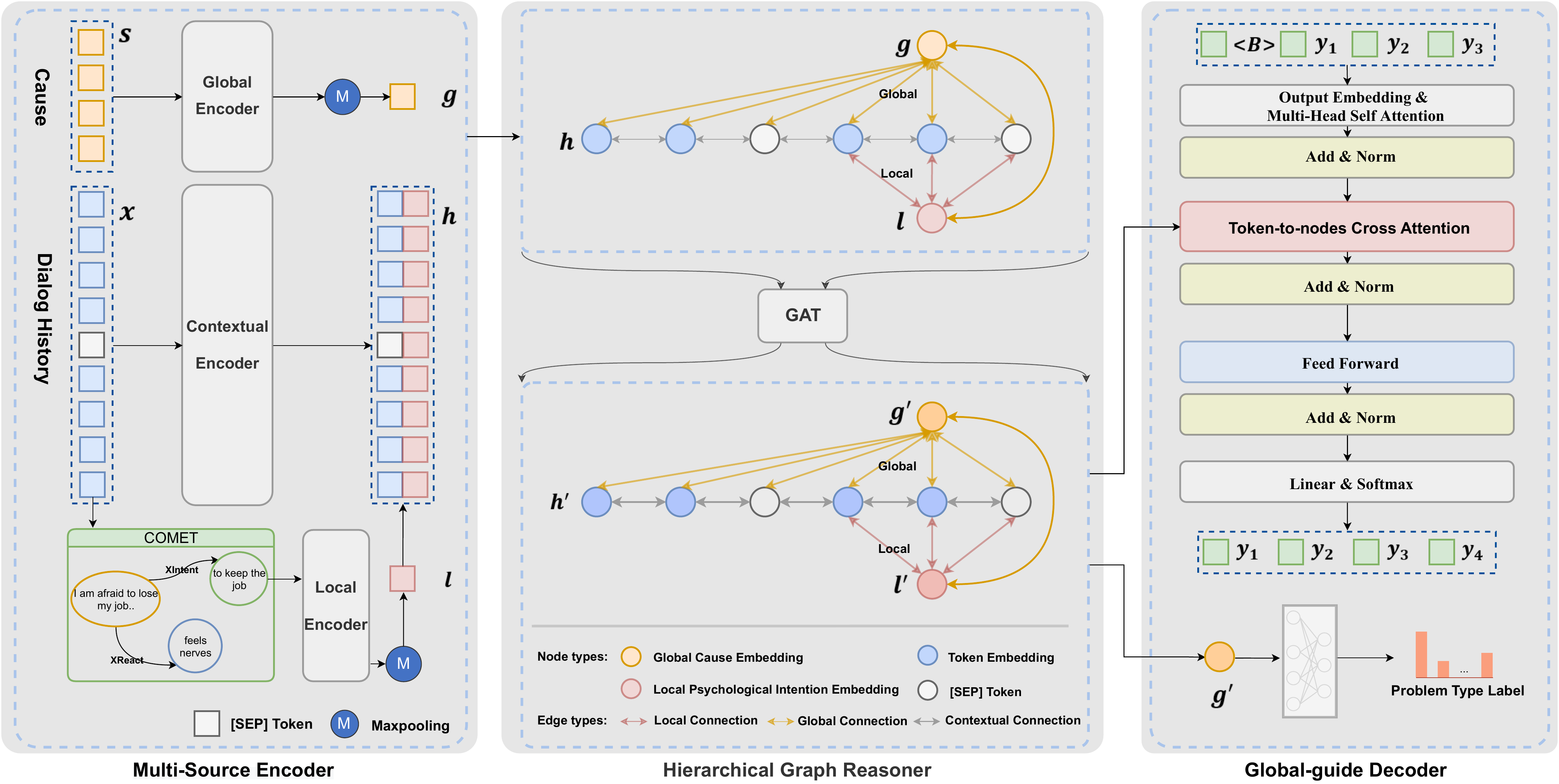}
	\caption{The overview of GLHG, which consists of Multi-source Encoder, Hierarchical Graph Reasoner and Global-guide Decoder. Green box indicates the pre-trained generative commonsense reasoning model COMET. }
	\label{fig:model}
\end{figure*}

\section{Approach}

As shown in Figure \ref{fig:model}, the proposed GLHG consists of three parts. Firstly, \textbf{Multi-source Encoder} captures the global cause information and local psychological intentions, as well as encoding the dialog history. Secondly, \textbf{Hierarchical Graph Reasoner} incorporates the different level representations for graph reasoning from the global-to-local perspective. Finally, \textbf{Global-guide Decoder} generates the supportive responses and monitors semantic information of the global cause by the novel designed training objective.

\subsection{Problem Formulation}

The problem formulation of our task could be formulated in the following. 
Given a dialog history $C = (u^1, u^2, \dots, u^{N-1})$, a set of $N-1$ utterances, where $u^n = (x^n_{1}, x^n_{2}, \dots, x^n_{M})$ that consists of 
${M}$ words, with $s = (s_1, s_2, \dots, s_P)$ being the corresponding global cause information of the help-seeker (describes the cause of the emotional problem), a sequence of $P$ words. GLHG generates a supportive response $Y$ conditioned on the help-seeker's global cause, local psychological intention and dialog history.

\subsection{Multi-Source Encoder}

Multi-Source Encoder considers three sources information, including dialog history, global cause and local intention. The contextual encoder, global encoder and local encoder have the same architecture with the encoder of BlenderBot \cite{Roller2021RecipesFB}, which is an open-domain conversational agent pre-trained on large-scale dialog corpora.

Firstly, representations of dialog history are obtained. Following the previous work \cite{DBLP:conf/acl/LiuZDSLYJH20,Roller2021RecipesFB}, we separate each utterance with [SEP] and prepend a special token [CLS] at the beginning of the dialog history to reconstruct the context input $C = ([CLS], u^1, [SEP], \dots, u^{N-1}, [SEP])$. To obtain the contextual representations, the contextual encoder ${\rm Enc}_{ctx}$ is utilized to encode each word $x$, leading to a series of context-aware hidden states $(\boldsymbol{h}_1, \boldsymbol{h}_2, \dots, \boldsymbol{h}_T)$, as:

\begin{equation}\label{equ:1}
\boldsymbol{h}_t = {\rm Enc}_{ctx}([CLS], x^1_{1}, \dots, x^1_{M}, [SEP], \dots, [SEP])
\end{equation} 
where $T$ is the max length of the input sequence, $\boldsymbol{h}_{t} \in \mathbb{R}^{d}$ is the $t$-th token in the $C$, and $d$ is the hidden size of the encoder.

Then, the global encoder ${\rm Enc}_{glo}$ obtains the global cause with the situation information, the max-pooling operation is performed to output representation of the whole sequence:
\begin{equation}\label{equ:2}
\boldsymbol{g} = {\rm {Max\verb|-|pooling}}({\rm Enc}_{glo}(s_{1}, \dots, s_{P}))
\end{equation} 

Finally, to capture the psychological intention of the help-seeker, COMET utilizes the last utterance of the help-seeker with the extra special relation tokens (e.g., ``$\langle$ head $\rangle$ I need another job to pay more bills $\langle$ /head $\rangle$ $\langle$ relation $\rangle$ \textit{xIntent} $\langle$ /relation $\rangle$ [GEN]”) and generates corresponding mental state inferences $( ms_1, \dots, ms_L )$ for \textit{XIntent} relation, where $L$ is the length of inferences sequence. Similarly, the module ${\rm Enc}_{loc}$ obtains the representation of the local intention by:

\begin{equation}\label{equ:3}
\boldsymbol{l} = {\rm {Max\verb|-|pooling}}({\rm Enc}_{loc}(ms_1, \dots, ms_L ))
\end{equation}
where $\boldsymbol{l} \in \mathbb{R}^{d}$. (Note, We only use \textit{XIntent} in this paper).

\subsection{Hierarchical Graph Reasoner}

The graph reasoner is proposed to organize the multi-source information and model the interactions between the global cause, local intention and dialog history, which enables the model to better build situational understanding of the conversation.

Following \cite{DBLP:journals/tnn/ScarselliGTHM09}, after obtaining the representations of the global cause, local psychological intentions and dialog history, we build a graph network \cite{DBLP:journals/tnn/ScarselliGTHM09} by connecting nodes with edges. And then we integrate relevant information for each graph node with a graph attention network (GAT) \cite{DBLP:journals/corr/abs-1710-10903}, which propagates features from other neighbourhood’s information to the current node and has the advantage of determining the importance and relevance between nodes. In this section, we first define the components of our graph $\mathcal{G} = (\mathcal{V}, \mathcal{E})$ and then introduce the reasoning process.
\paragraph{Vertices.} The representation of the global cause $\boldsymbol{g}$, each token $\boldsymbol{h}_t$ in the dialog history and local intention $\boldsymbol{l}$ are represented as the vertices in graph $\mathcal{G}$, which are initialized with the corresponding encoded feature vectors. We represent the vertex features as $\mathcal{V} = \{\boldsymbol{g}, \boldsymbol{{h}}_1, \dots, \boldsymbol{h}_T, \boldsymbol{l}\} = \{\boldsymbol{v}_1, \boldsymbol{v}_2, \dots, \boldsymbol{v}_{T+1}, \boldsymbol{v}_{T+2}\} \in \mathbb{R}^{(T+2) \times d}$.
Three types of vertices/nodes $\boldsymbol{v}_i \in \mathcal{V}$ include global cause embedding, token embedding and local psychological intention embedding.

\paragraph{Edges.} There are three types of edges in the graph.

\textit{{global connection}} The global connection is constructed for controlling the whole semantic information of the conversation, where node $\boldsymbol{g}$ is connected to all the other nodes to globally make the mutual interactions. 

\textit{{local connection}} The local connection is constructed for selecting most relevant context based solely on the last help-seeker's state. In this way, the node $\boldsymbol{l}$ should only be connected to tokens in the last utterance and global feature.

\textit{{contextual connection}} The contextual connection is constructed for maintaining the information flow between conversations, where node $i$ is linked to the contextual tokens to propagate features with its neighbourhood.

\paragraph{Graph Modeling.} By doing this, we describe the methodology to transform the sequentially encoded features using the graph attention network with the global connection, local connection and contextual connection. More specifically, we need to update the nodes of the three types $\boldsymbol{g} ^ {(k)}$, $\boldsymbol{{h}}_t ^ {(k)}$ and $\boldsymbol{l} ^ {(k)}$, which are presented to global node, token nodes and local node in the $k$-th layer of the graph, respectively. Firstly, $\boldsymbol{v}_1 ^ {(k+1)} = \boldsymbol{g} ^ {(k+1)}$, computed by aggregating neighbourhood information, as:

\begin{eqnarray}\label{equ:4}
\boldsymbol{v}_1 ^ {(k+1)} = \sigma\big(\sum_{j \in \mathcal{N}_g} \alpha_{j}^k \boldsymbol{W}_b^k {\boldsymbol{v}_j}^{(k)} \big) 
\end{eqnarray}
where $\mathcal{N}_g$ is the neighbors of the global node in the graph, $k = \{1, \cdots, K\}$, $\boldsymbol{W}_b^{k} \in \mathbb{R}^{d \times d}$ is the trainable weight matrix, and $\sigma$ represents the nonlinearity activation function.

The weight $\alpha_{j}^{k} = \mathcal{H}({\boldsymbol{v}}_{1}^{(k)}, {\boldsymbol{v}}_{j}^{(k)})$ in Equation (\ref{equ:4}) is calculated with an attention mechanism, which models the importance of each $\boldsymbol{v}_j^ {(k)}$ to $\boldsymbol{v}_1^ {(k)}$:
\begin{eqnarray}
\alpha_{j}^{k} = \frac{\exp(\mathcal{F}({\boldsymbol{v}}_{1}^{(k)}, {\boldsymbol{v}}_{j}^{(k)}))}{\sum_{j^\prime \in \mathcal{N}_g} \exp{(\mathcal{F}({\boldsymbol{v}}_{1}^{(k)}, {\boldsymbol{v}}_{j^\prime}^{(k)}))}} 
\end{eqnarray}
where $\mathcal{F}$ is an attention function.

Following \cite{DBLP:journals/corr/abs-1710-10903}, the attention function can be formulated as:
\begin{eqnarray}
\mathcal{F}({\boldsymbol{v}}_{1}^{(k)}, {\boldsymbol{v}}_{j}^{(k)}) = \operatorname{LeakyReLU}\left(\mathbf{a}^\top[\boldsymbol{W}_b^k{\boldsymbol{v}}_1^{(k)}\|\boldsymbol{W}_b^k{\boldsymbol{v}}_j^{(k)}]\right)
\end{eqnarray}
where  $\mathbf{a} \in \mathbb{R}^{2d}$ is the trainable weight matrix, LeakyReLU is a nonlinearity activation function, and $\cdot ^ T$ represents transposition and $\|$ is the concatenation operation.

Similarly, for token nodes and local node, the graph interaction update process can be formulated as:

\begin{eqnarray}\label{equ:5}
\boldsymbol{v}_i ^ {(k+1)} = \sigma\big(\sum_{j \in \mathcal{N}_t} \beta_{j}^k \boldsymbol{W}_d^k {\boldsymbol{v}_j}^{(k)} \big) \\
\boldsymbol{v}_{T+2} ^ {(k+1)} = \sigma\big(\sum_{j \in \mathcal{N}_l} \gamma_{j}^k \boldsymbol{W}_d^k {\boldsymbol{v}_j}^{(k)} \big) \label{equ:6}
\end{eqnarray}
where $\mathcal{N}_t$ and $\mathcal{N}_l$ are the neighbors of the token nodes and local node in the graph, respectively. $i = \{2, \cdots, (T+1)\}$,  $\beta_{j}^{k} = \mathcal{H}({\boldsymbol{v}}_{i}^{(k)}, {\boldsymbol{v}}_{j}^{(k)})$ and $\gamma_{j}^{k} = \mathcal{H}({\boldsymbol{v}}_{(T+2)}^{(k)}, {\boldsymbol{v}}_{j}^{(k)})$.

\subsection{Global-guide Decoder}
Global-guide Decoder generates the supportive responses with the updated features of nodes. In addition, a novel training objective monitors semantic information of the global cause.

\paragraph{Response Generation.}
The output $Y = (y_1, y_2, \dots, y_Z)$ consists of $Z$ words. Following the previous work \cite{Roller2021RecipesFB,DBLP:conf/acl/LiuZDSLYJH20}, the decoder aims at predicting the word probability distribution. At each decoding timestep $z$, it reads the word embedding ${\boldsymbol {W}}_{y<z}$ and the outputs of the graph reasoner ${\boldsymbol{v}}^{(K)}$ for decoding.

\begin{equation}\label{equ:decoder}
\boldsymbol{p}(y_z | \{y_1, \dots ,y_{z-1}\}, {\boldsymbol{v}}^{(K)}) = {\rm {Decoder}}({\boldsymbol {W}}_{y<z}, {\boldsymbol{v}}^{(K)})
\end{equation}

where ${\boldsymbol {W}}_{y<z}$ denotes the embeddings of the generated tokens. And we use the updated representations in the graph reasoner ${\boldsymbol{v}}^{(K)}$ to make a token-to-nodes cross attention, where token is in the decoder, nodes are in the graph reasoner.

\paragraph{Supervision of Global Semantic Information.}
Considering that the global cause controls the whole conversation flow, we design a novel task to monitor global semantic information with the category (e.g., the problem type of this conversation). The probability of the category can be calculated by leveraging the representation of the global node, as:

\begin{equation}\label{equ:global}
\boldsymbol{p}(o) = {\rm Softmax}({\rm MLP}({\boldsymbol{v}_1}^{(K)}))
\end{equation}

\subsection{Joint Training}
The standard negative log-likelihood loss and cross-entropy loss of the two tasks are optimized as:
\begin{equation} \label{equ:l1}
\mathcal{L}_{1}=-\sum_{z=1}^{Z} \log \boldsymbol{p}(y_z | \{y_1, \dots ,y_{z-1}\}, {\boldsymbol{v}}^{(K)})
\end{equation}

\begin{equation} \label{equ:l2}
\mathcal{L}_{2}=-\boldsymbol{ \hat{o}}\log \boldsymbol{p}(o)
\end{equation}
where $\boldsymbol{\hat{o}}$ is the true label (problem type) of the conversation.

We combine the above two loss functions as the training loss in a multi-task learning manner \cite{DBLP:journals/kbs/PengHYXX21} as:
\begin{equation} \label{equ:13}
\mathcal{L(\theta)}=\lambda_1\mathcal{L}_{1}+\lambda_2\mathcal{L}_{2}
\end{equation}
where $\theta$ is the all learnable parameters, and $\lambda_1$ and $\lambda_2$ are two hyper-parameters for controlling the weight of the rest tasks.

\begin{table*}[t]
	\centering
	\setlength\tabcolsep{12pt}
	\resizebox{\linewidth}{!}{
		\begin{tabular}{lcccccccc}
			\midrule
			\textbf{Model}   & \textbf{PPL}$\downarrow$	& \textbf{B-1}$\uparrow$	& \textbf{B-2}$\uparrow$	& \textbf{B-3}$\uparrow$	& \textbf{B-4}$\uparrow$	& \textbf{D-1}$\uparrow$  & \textbf{D-2}$\uparrow$ & \textbf{R-L}$\uparrow$ \\
			\midrule	
			Transformer$^*$ \cite{DBLP:conf/nips/VaswaniSPUJGKP17}    & 81.55                &  17.25     & 5.66  & 2.32 & 1.31  & 1.25  & 7.29  & 14.68             \\
			DialogueGCN$^*$ \cite{DBLP:conf/emnlp/GhosalMPCG19}    & 65.31                & 14.82     & 4.82  & 1.79 & 1.16  & 1.89  & 10.72  & 14.26              \\
			MoEL$^*$ \cite{DBLP:conf/emnlp/LinMSXF19}    						   & 62.93                & 16.02   & 5.02  & 1.90 & 1.14  & 2.71  & 14.92  & 14.21          \\
			MIME$^*$ \cite{DBLP:conf/emnlp/MajumderHPLGGMP20}    						   & 43.27               & 16.15   & 4.82  & 1.79 & 1.03  & 2.56  & 12.33  & 14.83          \\
			DialoGPT \cite{Zhang2020DIALOGPTL}   (117M) 						   & \textbf{15.51  }              & -  & 5.13  & - & -  & -  & -  & 15.26    \\
			\midrule
			BlenderBot-Joint$^*$ (90M)                       & 16.15   & 17.08  & 5.52  & 2.16 & 1.29  & 2.71  & 19.38  & 15.51    \\
			{\textbf{GLHG (ours)}} (92M) 	   & {15.67}   & \textbf{19.66}  & \textbf{7.57}  & \textbf{3.74} & \textbf{2.13}  &\textbf{ 3.50}  & \textbf{21.61}  & \textbf{16.37}       \\
			\midrule
	\end{tabular}}
	\caption{\label{tab:1} Performance of automatic evaluations. The best results are highlighted in \textbf{bold}. $^*$ indicates that the performance is reproduced.}
\end{table*}

\section{Experiments}
\subsection{Dataset \& Evaluation Metrics}
Emotional Supprot Conversation dataset contains 1,053 multi-turn dialogues with 31,410 utterances. Each conversation contains the cause information, dialog contextual, problem type. 
We make a statistic of the problem type in the ESConv, which includes 12 categories in total. We keep the train/test partition unchanged \cite{DBLP:conf/acl/LiuZDSLYJH20}.

\begin{table}[!]
	\setlength\tabcolsep{12pt}
	\centering
	\resizebox{\linewidth}{!}{
		\begin{tabular}{lccccc}
			\toprule
			\textbf{Comparisons} &
			\textbf{Aspects} &
			\textbf{Win} &
			\textbf{Lose} &
			\textbf{Tie} \\
			\midrule
			&  Flu. &\textbf{62.4$^\ddag$}& 23.1& 15.5\\
			& Ide. &\textbf{60.8$^\ddag$}& 21.5& 17.7\\
			GLHG vs. Transformer & Com.&\textbf{69.3$^\ddag$}& 14.9& 15.8\\
			& Sug.&\textbf{58.2$^\ddag$}&24.6&18.2\\
			& Ove.&\textbf{62.3$^\ddag$}&16.7&21.0\\
			\midrule
			&  Flu. &\textbf{60.1$^\ddag$} & 28.3& 11.6\\
			&  Ide. &\textbf{53.2$^\ddag$}&17.8&29.0\\
			GLHG vs. MIME  & Com.&\textbf{48.6$^\dag$}&35.7&15.7\\
			& Sug.&\textbf{49.2$^\dag$}&31.5&19.3\\
			& Ove.&\textbf{51.9$^\ddag$}&28.5&19.7\\
			\midrule
			&  Flu. &\textbf{56.3$^\dag$} & 38.6& 5.1\\
			& Ide. &\textbf{49.5$^\ddag$}&22.6&27.9\\
			GLHG vs. BlenderBot & Com.&\textbf{59.1$^\ddag$}&26.4&14.5\\
			& Sug.&\textbf{45.8$^\text{ }\text{ }$}&36.2&18.0\\
			& Ove.&\textbf{52.4$^\ddag$}&24.1&23.5\\
			\bottomrule
	\end{tabular}}
	\caption{
		Human evaluation results (\%). ${\dag}$,${\ddag}$ represent significant improvement with $p$-value $< 0.1/0.05$ respectively.
	}
	\label{table:h_results}
\end{table}

\paragraph{Automatic Evaluations.}
The automatic metrics include perplexity (PPL), BLEU-$n$ (B-$n$), ROUGE-L (R-L), Distinct-$1$ (D-$1$), and Distinct-$2$ (D-$2$) \cite{DBLP:conf/naacl/LiGBGD16}. Perplexity measures the high-level general quality of the generation model. D-$1$ / D-$2$ is the proportion of the distinct unigrams / bigrams in all the generated results to indicate the diversity.

\paragraph{Human A/B Evaluations.}
Given two models A and B in our case GLHG and baselines, respectively. Three workers are prompted to choose the better one (\textit{Win}) for each of the 128 sub-sampled test instances. The annotators can select a \textit{Tie} if the responses from both models are deemed equal. We adopt the same human evaluations with \cite{DBLP:conf/acl/LiuZDSLYJH20}: 1) Fluency (Flu.): which model’s responses are more fluent? 2) Identification (Ide.): which model is more helpful in identifying problems? 3) Comforting (Com.): which model is more skillful in comforting you? 4) Suggestion (Sug.): which model gives you more helpful suggestions? 5) Overall (Ove.): generally, which model’s emotional support do you prefer?

\subsection{Experimental Setting}
The implementation of GLHG is based on COMET \cite{DBLP:conf/acl/BosselutRSMCC19} \footnote{\small{~https://github.com/allenai/comet-atomic-2020}}  and Blenderbot \cite{Roller2021RecipesFB}. The AdamW optimizer \cite{Loshchilov2017FixingWD} with $ \beta_1 = 0.9$ and $\beta_2 = 0.99$ is used for training, with an initial learning rate of $3e-5$ and a linear warmup with $100$ steps. The mini-batch size is set to $16$ for training, and we use a batch size of 1 and a maximum of 40 decoding steps during inference. The max length of the input sequence is set to 128. The epoch is set to 5. $\lambda_1$ and $\lambda_2$ are set to $0.5$. All the models are trained on Tesla V-100 GPU and PyTorch. For a fair comparison, we concat the situation, extracted intention and the context as the inputs of all baselines. 

In the following, we provide some strong dialogue models: (1) \textbf{Transformer} \cite{DBLP:conf/nips/VaswaniSPUJGKP17}, the standard Transformer model. (2) \textbf{DialogueGCN} \cite{DBLP:conf/emnlp/GhosalMPCG19}, the output layer of DialogueGCN is modified with a decoder to generate response. (3) \textbf{MoEL} \cite{DBLP:conf/emnlp/LinMSXF19}, a Transformer-based model which softly combines the response representations from different transformer decoders. (4) \textbf{MIME} \cite{DBLP:conf/emnlp/MajumderHPLGGMP20}, another extension of Transformer model which mimics the emotion of the speaker. (5) \textbf{DialoGPT} \cite{Zhang2020DIALOGPTL}, which is a GPT-2-based model
pre-trained on large-scale dialog corpora. (6) \textbf{BlenderBot-Joint} \cite{Roller2021RecipesFB}, which is an open-domain conversational agent pre-trained with communication skills. Following \cite{DBLP:conf/acl/LiuZDSLYJH20}, we use the small version of BlenderBot.

\subsection{Experimental Results}
\paragraph{Automatic Evaluations.}
As depicted in Table \ref{tab:1}, our GLHG achieves promising results on almost all of the automatic metrics compared with state-of-the-art model BlenderBot-Joint and other baselines. GLHG reaches the lowest perplexity, which indicates that the overall quality of generated responses is higher than the baselines. In terms of Distinct-$n$, our model can generate diverse responses than other models. The remarkable improvements increase on all metrics, which indicate the effectiveness of global cause and local psychological intentions, as well as hierarchical graph modeling.

\paragraph{Human A/B Evaluations.}
The results in Table \ref{table:h_results} demonstrate that, with the hierarchical relationships with the global cause and local intentions, the responses from GLHG are more preferred than the responses from baselines by human judges. Compared with Transformer, our model has gained much on the all aspects, especially on Com. and Ove. metrics, which indicates that our model is better at comforting others and providing suggestions. Compared with MIME and BlenderBot, our GLHG also achieves remarkable advancements, which shows the strong ability of our model in emotional support. As for fluency metric, while GLHG does not significantly outperform BlenderBot, it still gains decent performance, which presents the advantage of GLHG in ability of expressing. In addition, we sample 300 examples and find over
80.67\% extracted intentions are logically consistent.

\begin{table*}[t]
	\setlength\tabcolsep{10pt}
	\begin{tabular}{p{1.8cm}|p{14.35cm}}
		\toprule
		{Cause}     & I have been put into sadness due to pressure from my employer who is threatening to down size the manpower at work. I am sad because my supervisor has mentioned to me that I am going to lose the jobs.                                                                                                                                                                    \\ \hline
		{Context}   & \begin{tabular}[c]{@{}l@{}}\textbf{Help-seeker}:  I am really sad and stressed up knowing that i am soon losing my job due to corona virus \\ for sure if I lose my job now i might go to depression because I have a family to take care of. \\
		
		\textbf{System}: Ok, don't worry friend. \\
		\textbf{Help-seeker}:  Are there any ways you know that could help me \textbf{\color{red}{convince my boss}} that it is not the right \\ time to cut down on manpower? \\
		\end{tabular} \\ \hline
		{Intention}     & To be able to \textbf{\color{red}{convince his boss}}.                                                                                                                                                                    \\ \hline
		{BlenderBot} & Well, I hope that you are able to get a job soon and move up in your life.                                                                                                                                     \\ \hline
		\textbf{GLHG}     & \textbf{\color{red}{I know it's hard}} to get people to understand, but you can \textbf{\color{red}{talk to your boss}} and see if he has any support.                                                                                                                                                   \\ \hline
		{Ground truth}      & Lots of people lose the job in this corona time.                                                                                        \\ \hline
		
		\hline
		\hline
		{Cause}     & Effects of the \textbf{\color{red}{pandemic}}.                                                                                                                                                                                                                                                                   \\ \hline
		{Context}   & \begin{tabular}[c]{@{}l@{}}\textbf{Help-seeker}: Hello I was wondering if I can discuss the effects of pandemic have on my mental health. \\ \textbf{System}: Hello. How has the pandemic affected your mental health?  \\ \textbf{Help-seeker}: Well, right now my city is undergoing a second wave and it was doing very well so far, but \\ in the past two weeks, have been a slow and steady surge of daily cases. Just feeling stress and fear ...\end{tabular} \\ \hline
		{Intention}     & To be \textbf{\color{red}{safe}}.                                                                                                                                                                    \\ \hline
		{BlenderBot} & That sounds hard. Is there any way you can talk about this with your doctor?                                                                                                                                                                                                                             \\ \hline
		\textbf{GLHG}      & I understand that, it has been very hard for me too. And it is better to \textbf{\color{red}{stay at home}}.    \\ \hline
		{Ground truth}      & I understand how hard living through this pandemic is. it is such crazy time! I also have felt fear and stress from the second wave of the pandemic.                                                                                                                                                    \\ \bottomrule
		\noalign{\vskip -1mm}

		
	\end{tabular}
	\caption{Generated responses from baselines and GLHG. \textbf{\color{red}{Red words}} indicate the critical part during dialog generation.}
	\label{compare}
\end{table*}

\begin{table}[!]	
	\centering
	\setlength\tabcolsep{10pt}
	\resizebox{0.86\linewidth}{!}{
		\begin{tabular}{lcccc}
			\toprule
			& {B-2 $\uparrow$}  & {B-4 $\uparrow$} & {D-1 $\uparrow$} & {R-L $\uparrow$} \\ \midrule
			~{GLHG} 	& \textbf{7.57}& \textbf{2.13} & {3.50} 	& \textbf{{16.37}}
			 \\	\midrule
			~{w/o~Local Intention}	& {6.58}	& {1.90}  & {2.95} & {16.03}\\
			~{w/o~Global Cause} & {6.01}	& {1.45} & {3.12}  & {15.61}\\
			~{w/o~$\mathcal{L}_{2}$ Loss}	& {6.15}	& {1.75}  & \textbf{3.58}  & {15.87}\\
			~{w/o~Graph Reasoner} 	& {5.83}	& {1.34}  & {2.96}  & {15.74}\\	\midrule
			BlenderBot-Joint		& {5.52}	& {1.29}  & {2.71} & {15.51}  
			 \\	\bottomrule
	\end{tabular}}
	\caption{\label{tab:ablation} The results of ablation study on model components. }
\end{table}

\subsection{Ablation Study}
To get better insight into our GLHG, we perform the ablation study.
Specifically, we design four variants of GLHG: 1) \textbf{w/o Local Intention}, the psychological intention is removed in Equation (\ref{equ:3}) and the local node is not involved in Equation (\ref{equ:6}); 2) \textbf{w/o~Global Cause}, the global information is neglected, which is similar with congfiguation (1); 3) \textbf{w/o $\mathcal{L}_{2}$ Loss}, the $\mathcal{L}_{2}$ loss is removed in Equation (\ref{equ:l2});
4) \textbf{w/o Graph Reasoner}, we replace the graph with the linear and concatenation operation. The results in Table \ref{tab:ablation} present that each component is beneficial to the final performance, which suggest information about both global cause and local intentions are necessary for understanding the help-seeker's emotional distress problem and mental state. Furthermore, the graph reasoner makes a contribution to the overall performance, which demonstrates the modeling of the hierarchical relationships has the potential to provide emotional support responses.

\subsection{Case Study}
Qualitatively, we have observed interesting examples from the SOTA model and GLHG in Table \ref{compare}. In the first case, it can be observed that the help-seeker is anxious about losing his job. According to the context, the local intention is implicitly expressed with the sentences \textit{any ways that could help me {convince my boss} that ...}, and our model extracts the descriptions of his intention \textit{to be able to convince his boss}. After constructing the relationships between the global cause and local intentions, GLHG outputs an appropriate and empathetic response. BlenderBot generates a fluent response but it does not answer the question of the help-seeker. Compared with the ground truth, it is interesting that our model can generate \textit{I know it's hard to} (comforting) and \textit{but you can talk to your boss} (suggestion) based on the captured cause and implied feelings of the help-seeker. Similarly, in the second case, GLHG generates \textit{it is better to stay at home} to provide suggestion for the intention \textit{to be safe}. And the whole conversation is about the pandemic (cause). Moreover, we also discover some special patterns that emerged from the generated responses, such as \textit{I have been in a similar situation} or \textit{I think it would be good to}. 
The phenomenon demonstrates that GLHG responds with comfort and expresses empathy as well as providing suggestions during the conversation, which is consistent with the aim of emotional support.

\section{Conclusion}
This paper concentrates on integrating the global cause and local psychological intentions into the dialog history for generating the supportive responses. We propose a Global-to-Local Hierarchical Graph network (GLHG) to capture the multi-source information and model hierarchical relationships from the global-to-local perspective. In addition, the novel training objective is designed to monitor the semantic information of the global cause. Experiments and analysis demonstrate that the GLHG has achieved promising performance and significantly improves the human evaluations. For the future work, adopting other mental states to emotional support conversations is still worth researching. 

%

\appendix

\bibliographystyle{named}
\bibliography{ijcai22}

\end{document}